\documentclass[sigconf]{acmart}

\AtBeginDocument{%
  }

\copyrightyear{2022}
\acmYear{2022}
\setcopyright{rightsretained}
\acmConference[ICIIT]{ICIIT}{February 24--26, 2023}{Danang, Viet Nam}
\acmBooktitle{International Conference on Intelligent Information Technology (ICIIT), February 24--26, 2023, Da Nang, Viet Nam}\acmDOI{10.1145/3549555.3549568}
\acmISBN{978-1-4503-9720-9/22/09}

\usepackage{amsmath}
\usepackage{wrapfig}

\begin{document}

\title{A Robust and Low Complexity Deep Learning Model \\ for Remote Sensing Image Classification}

\author{Cam Le$^{1,3}$}
\email{cam.levt123@hcmut.edu.vn}
\affiliation{
      \institution{HCMC University of Technology}
      \country{Viet Nam}
}

\author{Lam Pham}
\email{Lam.Pham@ait.ac.at}
\affiliation{
  \institution{Austrian Institute of Technology}
  \country{Austria}
}

\author{Nghia NVN}
\email{nghianguyenbkdn@gmail.com}
\affiliation{
  \institution{Pintel ltd.}
  \country{South Korea}
}

\author{Truong Nguyen$^{2,3}$} %
\email{truongnguyen@hcmut.edu.vn}
\affiliation{%
  \institution{HCMC University of Technology}
  \country{Viet Nam}
}

\author{Le Hong Trang$^{1,3}$} %
\email{lhtrang@hcmut.edu.vn}
\affiliation{
  \institution{HCMC University of Technology}
  \country{Viet Nam}
}

\thanks{1. Faculty of Computer Science and Engineering}
\thanks{2. Faculty of Electrical and Electronics Engineering}
\thanks{3. HCMC University of Technology (HCMUT) 268 Ly Thuong Kiet, District 10, Ho Chi Minh City, Viet Nam and Vietnam National University Ho Chi Minh City (VNU-HCM) Linh Trung Ward, Thu Duc City, Ho Chi Minh City, Vietnam.}

\renewcommand{\shortauthors}{Cam Le et al.}

\begin{abstract}
In this paper, we present a robust and low complexity deep learning model for Remote Sensing Image Classification (RSIC), the task of identifying the scene of a remote sensing image.
In particular, we firstly evaluate different low complexity and benchmark deep neural networks: MobileNetV1, MobileNetV2, NASNetMobile, and EfficientNetB0, which present the number of trainable parameters lower than 5 Million (M).
After indicating best network architecture, we further improve the network performance by applying attention schemes to multiple feature maps extracted from middle layers of the network.
To deal with the issue of increasing the model footprint as using attention schemes, we apply the quantization technique to satisfy the maximum of 20 MB memory occupation.
By conducting extensive experiments on the benchmark datasets NWPU-RESISC45, we achieve a robust and low-complexity model, which is very competitive to the state-of-the-art systems and potential for real-life applications on edge devices.
\end{abstract}

\maketitle


\section{INTRODUCTION}
\label{intro}

As the task of remote sensing image classification (RSIC) is considered as an important component in various real-life applications such as urban planning~\cite{thapa2009urban, netzband2007applied}, natural hazards detection~\cite{poursanidis2017remote, van2013remote}, environmental monitoring~\cite{van2013remote}, vegetation mapping or geospatial object detection~\cite{feng2015uav}, it has attracted much research attention in recent years.
Indeed, the research community, which focuses on RSIC tasks, has published diverse datasets of remote sensing image as well as proposed a wide range of classification models.
The most early dataset of remote sensing image, UCM~\cite{yang2010bag}, was publish in 2010.
In next years, various remote sensing image datasets were published such as WHU-RS19~\cite{Xia2010WHURS19} in 2012, NWPU VHR-10~\cite{cheng2014multi}, SAT6~\cite{basu2015deepsat} and RSSCN7~\cite{zou2015deep} in 2015, SIRI-WHU~\cite{zhao2015dirichlet} in 2016, AID~\cite{xia2017aid} and NWPU-RESISC45~\cite{cheng2017remote} in 2017, and OPTIMAL~\cite{wang2018scene} in 2018.
Among these datasets, NWPU-RESISC45~\cite{cheng2017remote} presents the largest number of 45 different image scenes and balanced number of 700 images per class.
Regarding RSIC systems, they can be separated into two approaches.
The first approach mainly focuses on image processing techniques and machine learning based classification.   
While the image processing techniques are used to extract distinct features from the original image data, the traditional machine learning methods are used to classify these extracted features into certain classes. 
Regarding image processing based feature extraction, a wide range of methods were proposed such as Texture Descriptors (TD), Color Histogram (CH), Scale-Invariant Feature Transformation (SIFT) \cite{yang2008comparing}, wavelet transformation with Gabor/Haar filters \cite{elmannai2013support, elmannai2016new}, bag-of-visual-words (BoVW) based techniques \cite{yang2010bag, sridharan2014bag}.
These methods make effort to transform the original image into a new and condense feature space, likely vector, which is suitable for traditional machine learning classification such as  Support Vector Machine (SVM)~\cite{yang2010bag, elmannai2016new}, K-means Clustering~\cite{zheng2008k}, or Decision Tree and Neural Network\cite{du2012multiple}.
In the second approach, RSIC research community focuses on deep learning based models, mainly using variants of Deep Convolutional Neural Network (DCNN) such as VGG \cite{ye2021lightweight}, ResNet \cite{shabbir2021satellite}, DenseNet \cite{tong2020channel}, EfficientNet\cite{zhang2020transfer}, or Transformer \cite{zhang2021trs}. 
To train these networks, there are 3 typical strategies\cite{nogueira2017towards}: direct training, fine tuning, and using DCNN as a feature extractor. 
While the first strategy directly trains a network architecture on a RSIC dataset~\cite{pham2022remote}, the other two methods make use of pre-trained models on large-scale image datasets to finetune~\cite{hu2015transferring, wang2022empirical, shabbir2021satellite} or extract features~\cite{minetto2019hydra, zhao2020remote, li2020augmentation, zhang2021best, li2021gated, pham2022remote} on a RSIC dataset (i.e. Leveraging a pre-trained models in these two training strategies is considered as the transfer learning technique).
As most of datasets of RSIC present a limitation of data compared with natural image datasets such as ImageNet~\cite{imagenet_dataset}, training a network from scratch shows high cost and present ineffective compared with fine tuning methods or using DCNN as a feature extractor.

Compare between two approaches, the second approach leveraging deep learning based systems proves robust and outperforms the traditional machine learning based approach~\cite{mehmood2022remote}. 
However, complicated deep neural networks in the second approach commonly presents very high model complexity which causes challenging for applying RSIC on edge devices. 
In this paper, we address the problems of those two approaches, aim to develop a robust and low-complexity deep learning model for RSIC task. We mainly contribute: 

\begin{enumerate}
\item Firstly, we evaluate and compare current benchmark and low-complexity network architectures: MobileNet, MobileNetV2, NASNetMobile, EfficientB0.
Our experimental results indicate that the EfficientNetB0 architecture using the transfer learning technique is more effective for RSIC task.\\

\item Secondly, we propose a Multihead attention based layer which is applied to multiple feature maps for improving EfficientNetB0 network performance. 
To deal with the issue of increasing model complexity using the attention layers, we apply the quantization technique to meet the requirement: The proposed model occupies a maximum of 20 MB which is potential for applying to a wide range of edge devices surveyed in \cite{sun2021mind}. \\ 

\item Finally, we evaluate our best model (EfficientNetB0 network architecture using the transfer learning technique, the proposed Multihead attention based layer, and the quantization technique) on the largest and benchmark dataset of NWPU-RESISC45~\cite{cheng2017remote}.  
The experimental results show that our proposed RSIC system is competitive to the state of the art, but presents significantly lower model footprint. 
\end{enumerate}

\section{Background}

As our proposed deep learning model leverages the parameter-based transfer learning technique and attention schemes, the background of these two techniques is comprehensively presented below.

\subsection{The parameter-based transfer learning applied for deep neural network}
\label{transfer}
Humans can be aware that it is easy to transfer knowledge from one domain or task to another. 
For an instance, it will be easier for a person to learn a second programming language if he/she had experience on a programming language before.
In other words, a person can encounter a new task without starting from scratch by leveraging previous experience to learn and adapt to a new task.
Inspired by the human capability to transfer knowledge, the machine learning research community has recently focused on the transfer learning techniques and made effort to apply on the computers~\cite{survey_transfer, survey_transfer_02}.

In this paper, we apply the parameter-based transfer learning technique, which is very popular and effective for deep neural network network~\cite{pires2019convolutional}. 
Given a model of neural network architecture, we firstly define the term of `pre-trained model': A model was trained on a particular large-scale dataset for a certain task in advance, referred to as the up-stream task.
Then, transfer learning is a term that points out the action of applying the pre-trained model for a new task but related in some aspect of the up-stream task. 
The new task is referred to as the down-stream task.
Commonly, the up-stream task is more challenging than the down-stream task (e.g., more objects in tasks of object detection or more categories in classification tasks) and the dataset used in the down-stream task is normally smaller or more specific than the large-scale and general dataset for the up-stream task. 
The idea and advantages behind the parameter-based transfer learning technique for deep neural network is that utilizing the information gained while solving a challenging up-stream task (i.e. The trainable parameters and the network architecture of the pre-trained model) may not only save time but also enhance the performance on a more simple down-stream task.
Regarding the mathematical perspective behind the classification task and deep neural network based model in this paper, it is basically an optimization task which makes gradient descent find the minimum point.
Therefore, the starting point of gradient is a very important factor.
Indeed, if the starting point of gradient is near the global optimum point, it significantly helps to save the training time as well as avoid the gradient to converse at unexpected local optimization points.
By applying the parameter-based transfer learning technique, the distribution of trainable parameters, which is reused from a pre-trained model on an up-stream task, is likely to be near the golden distribution of trainable parameters in a down-stream task rather than random initialization. 
As the start distribution of trainable parameters is likely same as the golden distribution of trainable parameters, the gradient feasibly converse at very near the global optimal point. 

In this paper, we aim to classify remote sensing image into sentiment categories, which is considered as the task of remote sensing image classification (RSIC).
As we leverage the parameter-based transfer learning technique, our task of RSIC is referred to as the down-stream task.
To solve our down-stream task of RSIC, we there need to define the up-stream task of image classification as well as indicate a pre-trained model with a large-scale dataset.
As ImageNet is considered as the benchmark dataset~\cite{Imagenet} to evaluate a wide range of network architectures on the task of image classification, published pre-trained models on ImageNet from Keras library~\cite{keras_app} are considered as the up-stream tasks and leveraged for our down-stream task of RSIC.
 
\subsection{Attention schemes in computer vision}
\label{attention}

Humans can easily find the important regions in an image.
In other words, there are some regions on a image containing specific and distinct features which help humans distinguish from other images. 
This inspires the computer vision research community focuses on attention mechanisms which help deep learning models know and learn which valuable features.  
An attention mechanism can be formulated by a function $g(\mathbf{X})$ where $\textbf{X}$ is the input feature map and $g(\mathbf{X})$ represents a way to create the guidance based on the importance of input feature map $\mathbf{X}$.
In other words, the output of $g(\mathbf{X})$ is attention weights which present which region of the input feature map is more important.
The attention weights are then element-wise multiplied with the input feature map $\textbf{X}$~\cite{hu2018squeeze, woo2018cbam} as described by Eq.\ref{eq:att}
\begin{align}
   \label{eq:att}
   f(\mathbf{X}) = g(\mathbf{X}) \odot \mathbf{X}
\end{align}
where $f()$ is the attention layer applied on the input feature map $\mathbf{X}$ to generate a new feature map which better presents distinct features, but still retain the original feature map size.

The current attention mechanisms applied to the computer vision research field and deep learning models can be divided into some main groups described in detail below.

\textbf{Squeeze-and-excitation networks (SE)}~\cite{hu2018squeeze}. It is a channel-based attention mechanism, SE layer focuses on the particular features on the channel dimension. Moreover, SE uses global average pooling (GAP) before feeding to a multi layers perceptron neural network with a sigmoid function at the last layer. Then, it further applies a channel-wise multiplication between the input feature map $\mathbf{X}$ and the output of activation layer.
The SE is formulated by Eq.~\ref{eq:se}
\begin{align}
   \label{eq:se}
   f_{SE} = g(mlp(GAP(\mathbf{X}))) \odot \mathbf{X}
\end{align}
where ${g()}$ is sigmoid function, $mlp()$ stands for multi-layers perceptron neural network and GAP is a channel wise global average pooling layer.


\textbf{Channel attention (CA) layer:} CA layer is a variant of SE and it is also a channel-based attention method which has been popularly used in convolutional neural networks~\cite{guo2019global, Zhao_2022}. Similar to SE layer, the idea behind the channel attention layer is guiding the model to focus on some particular features on the channel. But it seems to be more powerful as it utilize information from both global max and average pooling layer.


In particular, given three-dimensional input feature map $\mathbf{X}\in R^{W\times H\times C}$ where $W, H,$ and $C$ are width, height, and channel dimensions, the channel attention (CA) applied to the feature map $\mathbf{X}$ can be formulated by:
\begin{align}
   \label{eq:ca}
   f_{CA} = g(mlp(GAP(\mathbf{X})) +  mlp(GMP(\mathbf{X})))  \odot \mathbf{X}
\end{align}
where ${g()}$ is sigmoid function, $mlp()$ is a sharing neural layer (e.g. normally use multi-layers perceptron). 
GAP and GMP are global average pooling and global max pooling of channel wise, respectively.

\textbf{Spatial attention (SA) layer:} enables the deep neural network focus on distinct features on both width and height dimensions rather than the channel dimension as CA or SE mechanisms.
As focusing on the spatial features on width and height dimensions, the channel dimension of a three-dimensional input feature maps $\mathbf{X}$ is firstly reduced by using average pooling and max pooling, create two-dimension feature maps of $\mathbf{X_{A}, X_{M}}{\mathnormal{\in R^{W\times H}}}$, respectively.
Then, a network layer (e.g., normally a convolutional layer), described by $conv()$ is applied and followed by a Sigmoid function.
The SA layer is formulated as Eq.~\ref{eq:sam}
\begin{align}
   \label{eq:sam}
   f_{SA} = g(conv( [\mathbf{X_{A}}, \mathbf{X_{M}}] ) ) \odot \mathbf{X}
\end{align}
where $g()$ is sigmoid function, $conv()$ represents for a convolutional layer.
 
\textbf{Convolutional Block Attention Module (CBAM):} While SE/CA and SA mechanisms only focus on either channel features or spatial features, CBAM~\cite{woo2018cbam}, combines both these attention methods, creates a robust guidance for network to process important regions of a certain feature map.
This attention mechanism can be described by formulas: Eq.\ref{eq:cbam1} and Eq.\ref{eq:cbam2}:
\begin{align}
   \label{eq:cbam1}
   \mathbf{X'} =  f_{CA}(\mathbf{X}) \\
   \label{eq:cbam2}
   f_{CBAM} =  f_{SA}(\mathbf{X'}) 
\end{align}

\textbf{Multihead self attention (MSA):} Unlike above methods which make effort to enhance important regions of a feature map, this attention scheme \cite{vaswani2017attention} helps to indicates the similarity score, the dependency between regions in the feature map.
In other worlds, Multihead self attention is effective to represent the relation between two regions of a feature map which are closed or far from each others. 
Regarding the mathematical intuition behind the Multihead self attention, each attention head can be described as mapping a query ($\mathbf{Q}$) and a set of key($\mathbf{K}$)-value($\mathbf{V}$) pairs to an output, where $\mathbf{Q}$, $\mathbf{K}$, $\mathbf{V}$ obtained through a linear transformation of the input feature map $\mathbf{X}$ as shown in Eq.\ref{eq:mha_q}, \ref{eq:mha_k}, and \ref{eq:mha_v}.
Then, the output of an attention head can be calculated using Eq.~\ref{eq:one_head}. 
\begin{align}
   \label{eq:mha_q}
   \mathbf{Q} =  \mathbf{X} \cdot \mathbf{W_{q}} \\
   \label{eq:mha_k}
   \mathbf{K} =  \mathbf{X} \cdot \mathbf{W_{k}} \\
   \label{eq:mha_v}
   \mathbf{V} =  \mathbf{X} \cdot \mathbf{W_{v}} \\
   g_{n} =  softmax(\frac{\mathbf{Q}\mathbf{K^{T}}}{\sqrt{\mathbf{d_{k}}}})\mathbf{V} 
  \label{eq:one_head}
\end{align}
where $g_{n}$ is the $n^{th}$ attention head, $\mathbf{W_{q}, W_{k}, W_{v}}$ are weight matrices, $\mathbf{K^{T}}$ is the transpose of $\mathbf{K}$ and $\mathbf{d_{k}}$ is the number of key dimension which is one of the dimension of the weight matrices.

As each attention head learns a different set of weight matrices, they will be different from each others. 
Therefore, when joining many self attention heads together followed by a linear transformation or an addition operation as an ensemble of multiple heads, it forms a Multihead self attention layer which helps to learn an input feature map better.
A Multihead self attention layer with $N$ heads which is applied on the input feature map $\mathbf{X}$ is described by
\begin{align}
    \label{eq:mha_head}
     f_{MA}=  \sum_{n=1}^{N} g_{n}   \odot \mathbf{X}
\end{align}

\section{Proposed deep learning based system for RSIC task}
\label{framework}

\begin{figure}[t]
    \centering
    \includegraphics[width=1.0\linewidth]{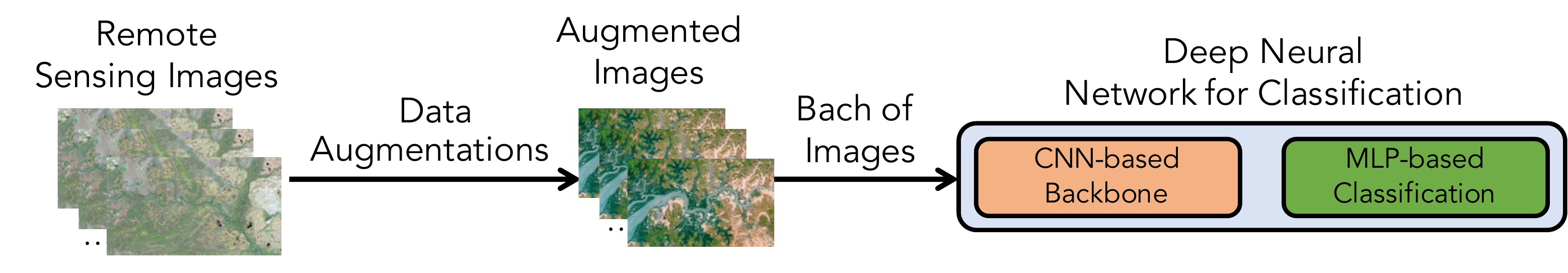}
	\caption{The high level architecture of proposed RSIC system}
    \label{fig:CAM_high}
\end{figure}

Overall, the high-level architecture of our proposed deep learning based system for RSIC task is presented in Figure~\ref{fig:CAM_high}. 
As Figure~\ref{fig:CAM_high} shows, the proposed RSIC system is separated into two main parts: data augmentation methods and a deep neural network for classification.

\subsection{Data augmentation methods}
\label{augmentation}

In this paper, we apply five data augmentation methods: Image Rotation (IR)~\cite{rotation_aug}, Random Cropping (RC)~\cite{rotation_aug}, Random Erasing (RE)~\cite{spec_crop}, Random Noise Addition (RNA), and Mixup (Mi)~\cite{mixup1, mixup2} to the remote sensing image input data.
As Random Cropping (RC)~\cite{rotation_aug}, Random Erasing (RE)~\cite{spec_crop}, Random Noise Addition (RNA), and Mixup (Mi)~\cite{mixup1, mixup2} are used on batches of images during the training process, they are referred to as the online data augmentation methods. 
Meanwhile, Image Rotation (IR)~\cite{rotation_aug} is referred to as the offline data augmentation as this method is applied on the original dataset before the training process.

\begin{figure}[t]
    \centering
    \includegraphics[width=1.0\linewidth]{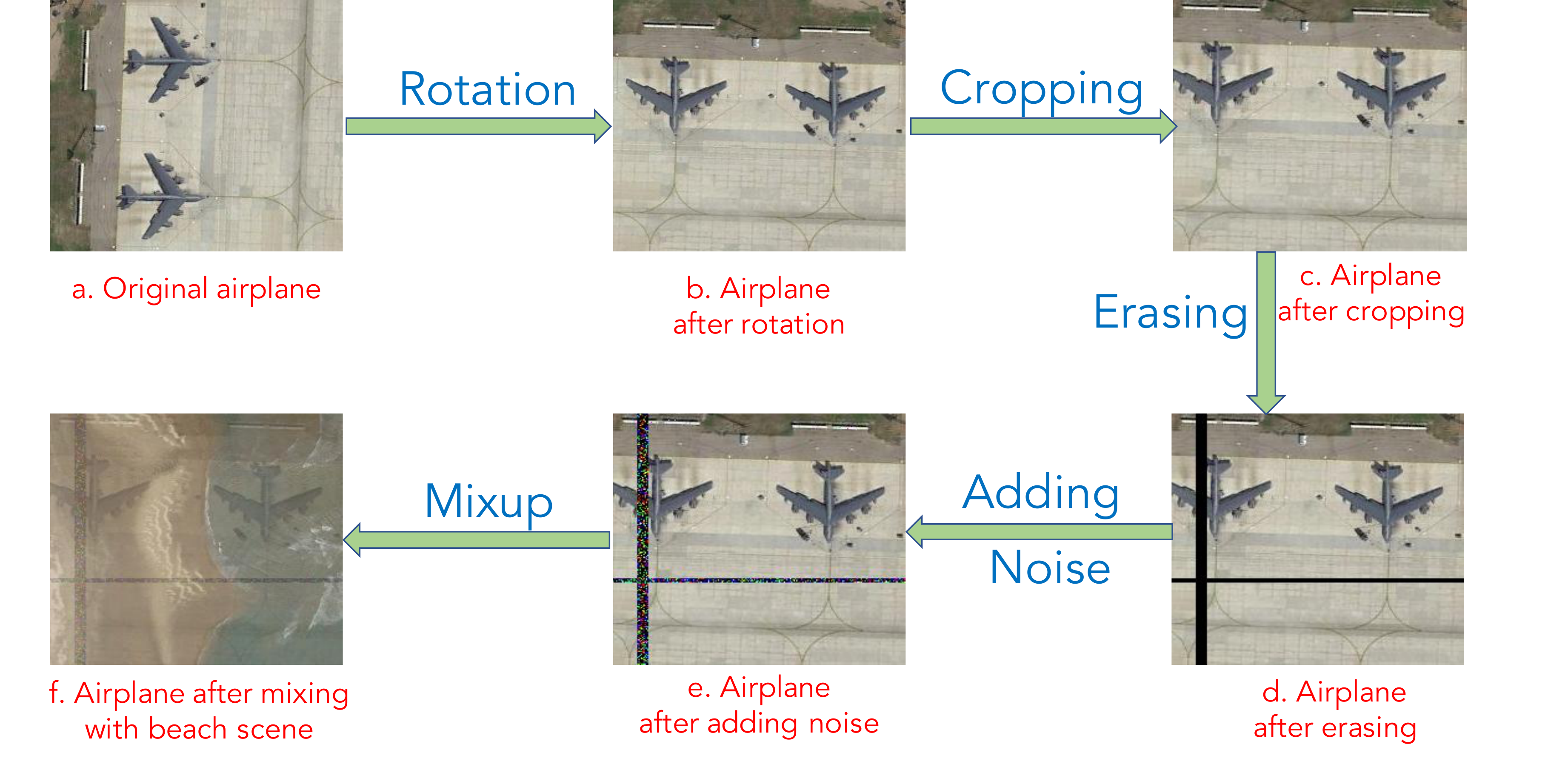}
	\caption{Data augmentation methods: Rotation, Random Cropping, Random Erasing, Adding Noise, and Mixup in the order.}
    \label{fig:CAM_aug}
\end{figure}

Initially, all images in the original dataset are rotated using three different angles: 90, 180, and 270, respectively.
This data augmentation method is referred to as Image Rotation (IR) and an example of IR method with an angle of 90 degree is shown in Figure~\ref{fig:CAM_aug} (b).
As three angles mentioned are used, we obtain a new dataset which is four times larger than the original dataset (i.e. the original images and three new images generated by Image Rotation method with three angles).
Next, batches of 60 images are randomly selected from the new dataset.  
For each batch, we apply Random Cropping (RC)~\cite{rotation_aug}, Random Erasing (RE)~\cite{spec_crop}, Random Noise Addition (RNA), and Mixup (Mi)~\cite{mixup1, mixup2} methods, respectively.
Firstly, images in a batch are randomly cropped with a reduction of 10 pixels on both of width and height dimensions as shown in Figure~\ref{fig:CAM_aug} (c) (i.e., The channel dimension is retained), referred to as Random Cropping (RC).
Next, on both width and height dimensions of each image, 20 random and continuous pixels are erased as shown in Figure~\ref{fig:CAM_aug} (d), referred to as Random Erasing (RE). 
The cropped and erased images are then added by a random noise which is generated from Gaussian distribution as shown in Figure~\ref{fig:CAM_aug} (e), referred to as Random Noise Addition (RNA). 
Finally, the images are mixed together with random ratios as shown in Figure~\ref{fig:CAM_aug} (f), referred to as Mixup (Mi).
As both Uniform and Beta distributions are used to generate the mixup ratios as well as we use both the original image and the new mixup images, the batch size increases three times from 60 to 180 images.

\subsection{Apply the transfer learning technique for our proposed deep neural network classification}
\label{neural}

As Figure~\ref{fig:CAM_high} shows, our proposed deep learning model for classification can be separated into two main parts: The convolutional neural network based backbone (CNN based backbone) and the multilayer perceptron based classification (MLP-based classification).
While the CNN based backbone helps to transfer the input images to condensed feature maps, the MLP based classification classifies these condensed feature maps into certain categories.

\begin{figure}[t]
    \centering
    \includegraphics[width=1.0\linewidth]{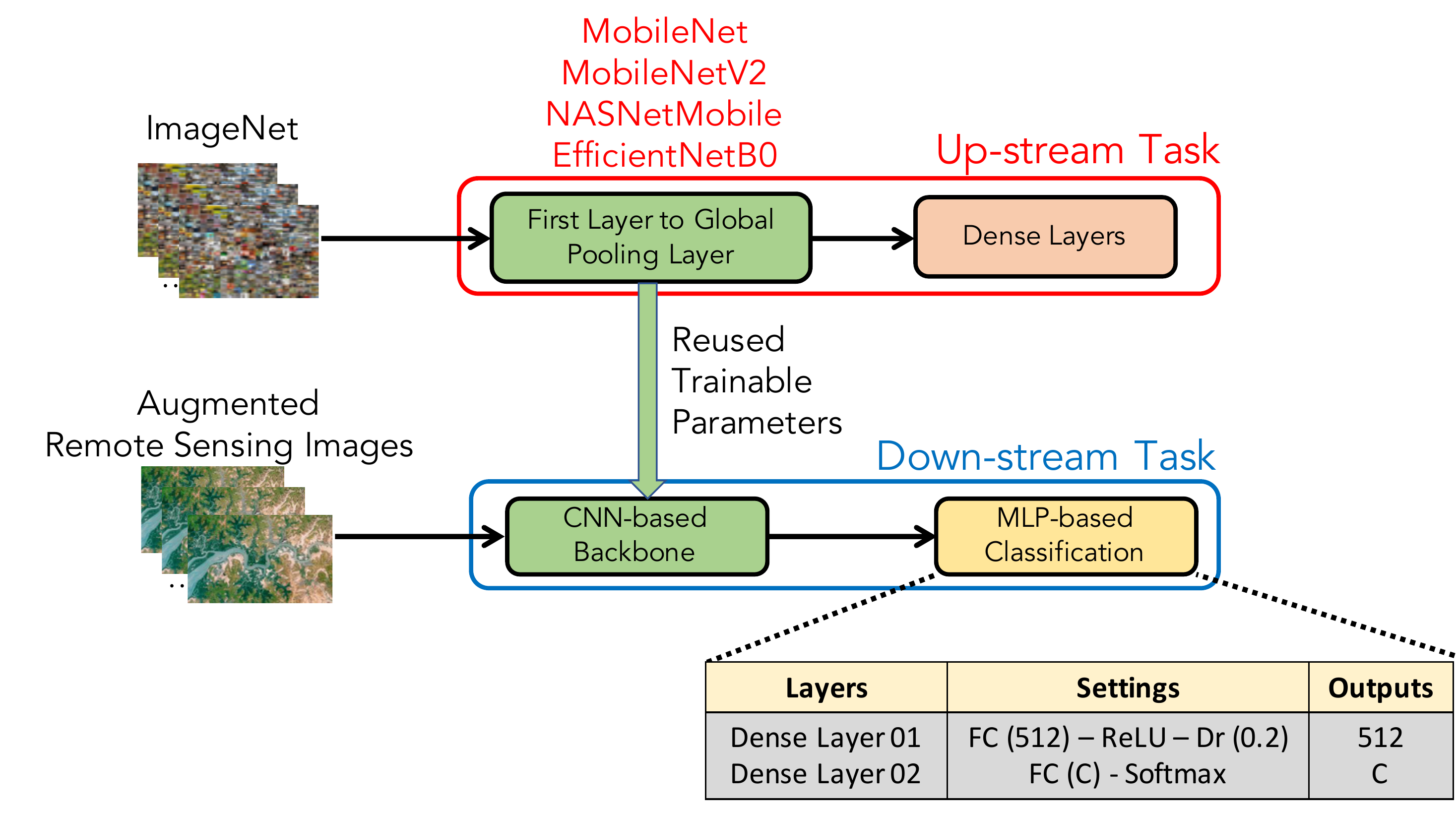}
	\caption{Apply the transfer learning technique for the proposed deep neural network classification}
    \label{fig:CAM_transfer}
\end{figure}

To indicate which CNN based backbone is effective for RSIC task, we evaluate different benchmark deep neural network architectures which are available in Keras library~\cite{keras_app}.
As we aim to achieve a low-complexity model for RSIC which is lower than 5 M of trainable parameters, only four network architectures of MobileNetV1, MobileNetV2, NASNetMobile, and EfficientNetB0 from Keras library~\cite{keras_app} are evaluated.
To leverage these network architectures, we apply the parameter-based transfer learning technique which is mentioned in Section~\ref{transfer}.
The transfer learning process is mainly described in Figure~\ref{fig:CAM_transfer}.
In particular, the benchmark networks of MobileNetV1, MobileNetV2, NASNetMobile, and EfficientNetB0 as described in the higher part of Figure~\ref{fig:CAM_transfer} are firstly trained with the large scale dataset of ImageNet, referred to as the up-stream task.
Next, only the first layer to the global pooling layer of these pre-trained models are re-used and considered as the CNN-based backbone.
The CNN based backbone is then connected with a MLP-based classification, create an end-to-end neural network model for the down-stream task on the target RSIC dataset.

The MLP-based classification as shown in the bottom-right part in Figure~\ref{fig:CAM_transfer} performs two dense layers (Dense Layer 01 and 02).
The first dense layer comprises one fully connected layer (FC(channel number=512)) followed by a rectified linear unit (ReLU)~\cite{relu} and a dropout (Dr(drop ratio))~\cite{dropout}.
Meanwhile, the second dense layer uses Softmax layer after the fully connected layer.
Notably, the number of channels at the second fully connected layer is set to $C$ that presents the number of categories in the target RSIC dataset.

\subsection{Apply attention schemes and explore multiple feature maps to further improve the proposed RSIC system}
\label{app_att}
\begin{figure}[t]
    \centering
    \includegraphics[width=1.0\linewidth]{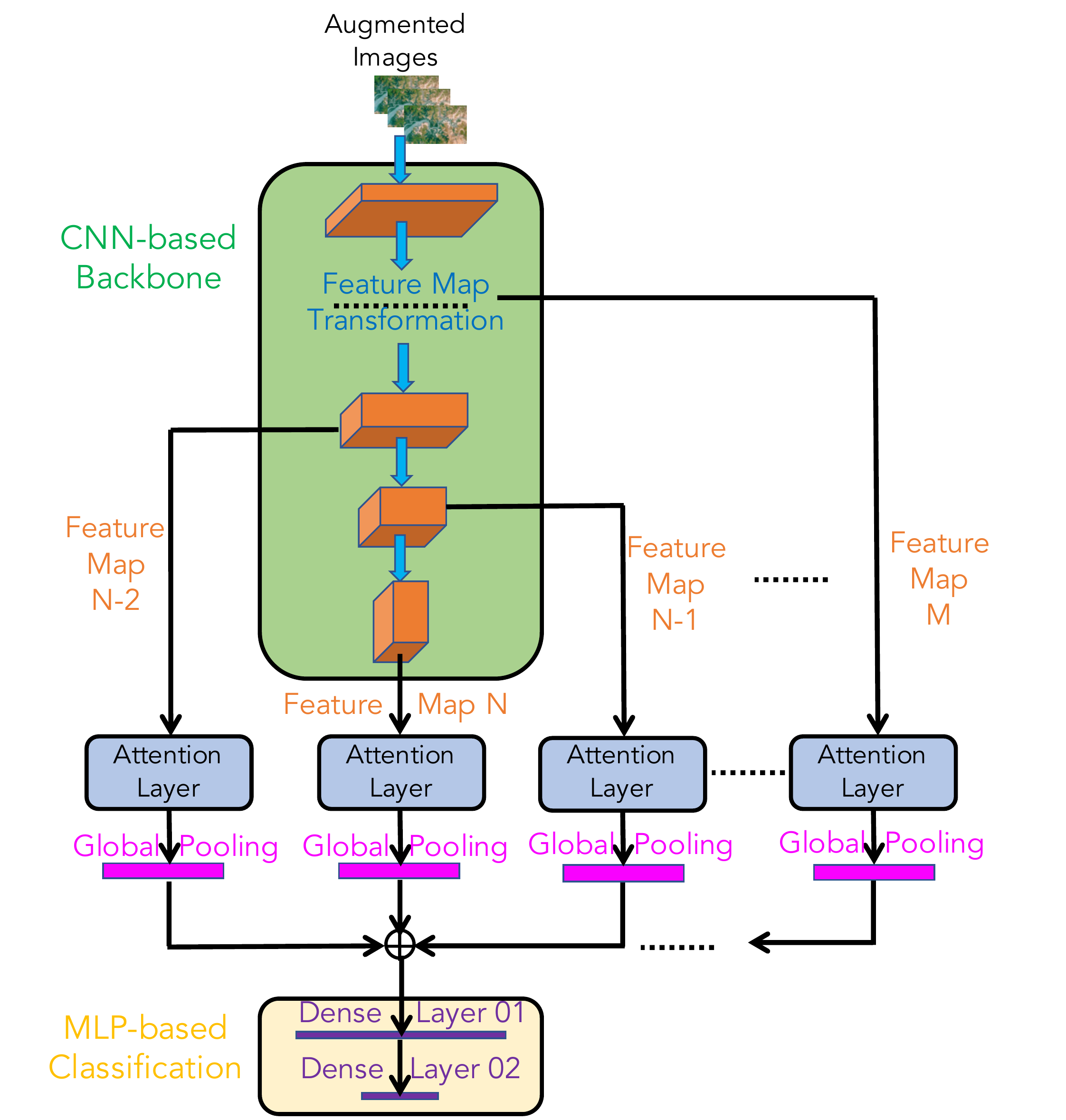}
	\caption{Apply attentions schemes to further improve the proposed deep neural network classification}
    \label{fig:CAM_apply_att}
\end{figure}
\begin{figure}[t]
    \centering
    \includegraphics[width=1.0\linewidth]{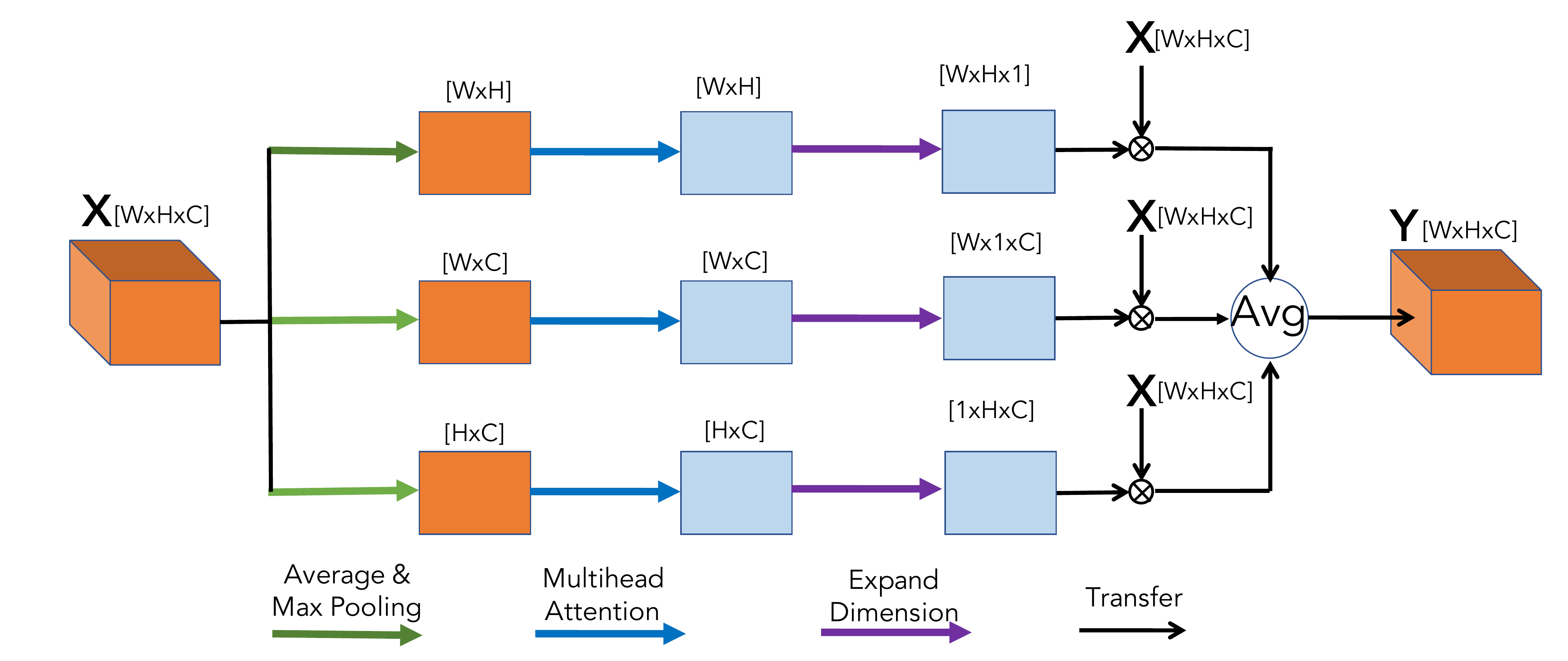}
	\caption{Proposed Multihead attention based layer.}
    \label{fig:CAM_pro_att}
\end{figure}

To further improve the proposed RSIC system, we apply different attention schemes mentioned in Section~\ref{attention} to feature maps extracted from middle layers of the CNN based backbone as shown in Figure~\ref{fig:CAM_apply_att}.
The feature maps are the final outputs of convolutional blocks of the CNN based backbone.
For an example, EfficientB0 based backbone presents 7 convolutional blocks, namely block 1 to block 7~\cite{tan2019efficientnet}.
Regarding the attention layer used in Figure~\ref{fig:CAM_apply_att}, we evaluate three types of attention schemes: SE, CBAM, and Multihead attention.
The first two attention layers of SE and CBAM are constructed basing on the formulations mentioned Section~\ref{attention}.
For the Multihead attention scheme, we propose a Multihead attention based layer as shown in Figure~\ref{fig:CAM_pro_att} which addresses drawbacks of SE or SA (i.e. SE and SA focus on either channel feature or spatial feature).
In particular, given an input feature map $\mathbf{X}$ with a size of [W$\times$H$\times$C] where $W$, $H$, and $C$ presents width, height, and channel dimensions, we reduce the size of feature map $\mathbf{X}$ across three dimensions using both max and average pooling layers.
We then generate 3 matrices: [W$\times$ H], [H$\times$ C], [W$\times$C]. 
Next we feed all generated matrices into thee Multihead attention to obtain attention score matrices. 
Then, we reshape the attention score matrices into the sizes of [W$\times$H$\times$1], [1$\times$H$\times$C],[W$\times$1$\times$C] respectively and element-wise multiply each of them with the input feature map $\mathbf{X}$. 
Finally, we conduct an average of three results of multiplications, generate the output tensor $\mathbf{Y}$ with the size of [W$\times$H$\times$C] which is same size as the input feature map $\mathbf{X}$.
Notably, we set the number of heads to 32 and set the key dimension to 8 for each Multihead attention.
By applying our proposed Multihead attention based layer, both channel feature (feature maps with sizes of [H$\times$ C], [W$\times$C]) and spatial feature (feature map with size of [W$\times$ H]) are focused, which help the network learn distinct features from the input feature map better.

As SE, CBAM, or our proposed Multihead attention base layers only transforms an input feature map $\mathbf{X}$ to an output feature map $\mathbf{Y}$ and retains the size of the input feature maps, we then apply a global average pooling layer after each attention layer to scale down the feature maps to vectors which is suitable for the MLP-based classifier for classification as shown in Figure~\ref{fig:CAM_transfer}.
  
\section{Experimental results and discussion}

\subsection{Dataset}
\label{dataset}

In this paper, we evaluate the benchmark dataset NWPU-RESISC45\cite{nwpu_dataset}.
NWPU-RESISC45 dataset was collected from more than 100 countries and regions in the world, consists of 31,500 remote sensing images. 
The remote sensing images are separated into 45 scene classes, each of which  700 images in RGB color format with resolution of $256\times256\times3$
To compare with the state-of-the-art systems, we obey the original settings mentioned in~\cite{nwpu_dataset}.
We then split the NWPU-RESISC45 dataset into Training and Testing subsets with two different ratios: 20\%-80\%  and 10\%-90\%, respectively.


\subsection{Evaluation Metrics}
To compare with the state-of-the-art systems, Accuracy (Acc.\%) is used as the main metric, which was proposed in almost benchmark datasets of AID\cite{aid_dataset}, UCM\cite{yang2010bag}, or NWPU-RESISC45\cite{nwpu_dataset}.  

\subsection{Model implementation and settings}
\label{setting}
As the data augmentation method of Mixup~\cite{mixup2} is applied, the ground truth are not in one-hot encoding format. 
We therefore apply Kullback-Leibler divergence (KL) loss~\cite{kl_loss} instead of Entropy loss.
\begin{align}
   \label{eq:kl_loss}
   Loss_{KL}(\Theta) = \sum_{n=1}^{N}\mathbf{y}_{n}\log \left\{ \frac{\mathbf{y}_{n}}{\mathbf{\hat{y}}_{n}} \right\}  +  \frac{\lambda}{2}||\Theta||_{2}^{2} \,,
\end{align}
where  \(\Theta\) presents trainable parameters, the constant \(\lambda\) is empirically set to $0.0001$, the batch size $N$ is set to 60, $\mathbf{y_{i}}$ and $\mathbf{\hat{y}_{i}}$ denote expected and predicted results, respectively.
We construct proposed deep learning networks with Tensorflow framework using
Adam~\cite{adam} for optimization.
The training and evaluating processes are conducted on two GPU Titan RTX 24GB.
The training process is stopped after 60 epoches. 
While the first 50 epoches uses the learning rate of 0.0001 and all data augmentation
methods mentioned in Section~\ref{augmentation}, the remaining 10 epoches uses the lower learning rate of 0.000001 with only the offline Random Rotation data augmentation method

\subsection{Experimental results}
\begin{table}[t]
    \caption{Performance (Acc.\%) of EfficientNetB0 with the proposed Multihead attention applied for feature maps extracted from different convolutional blocks on the benchmark NWPU-RESISC45 dataset using 20\%-80\% splitting settings.} 
    \centering
    \scalebox{1.0}{
    \begin{tabular}{|c|c|c|} 
        \hline 
        \textbf{Convolutinal Blocks}  &\textbf{Accuracy (\%)}        &\textbf{Parameters (M)} \\
  	    \hline         
        \textbf{Block 7}       &93.1 &6.0\\
        \textbf{Blocks 6 to 7}  &93.0 &7.5\\
        \textbf{Blocks 5 to 7}  &93.8 &9.4\\
        \textbf{Blocks 4 to 7}  &92.8 &11.2\\
        \textbf{Blocks 3 to 7}  &92.5 &13.3\\
         \hline 
      \end{tabular}
    }
    \label{table:res03} 
\end{table}
\begin{table}[t]
    \caption{Performance comparison among benchmark network architectures, with the transfer learning technique and without attention scheme, on the benchmark NWPU-RESISC45 dataset using 20\%-80\% splitting settings.} 
    \centering
    \scalebox{1.0}{
    \begin{tabular}{|l|c|c|} 
        \hline 
        \textbf{Network}  &\textbf{Accuracy (\%)}        &\textbf{Parameters (M)} \\
  	    \hline         
        \textbf{MobileNet} &90.2 &3.7\\
        \textbf{MobileNetV2} &90.9 &2.9\\
        \textbf{NASNetMobile}  &91.7 &4.8\\
        \textbf{EfficientNetB0}  &92.0 &4.6\\
         \hline 
      \end{tabular}
    }
    \label{table:res01} 
\end{table}

According to the results shown in Table \ref{table:res01}, the proposed RSIC system using the transfer learning technique and EfficientNetB0 and NASNetMobile based architectures are competitive and outperform MobileNet and MobileNetV2.
As EfficientNetB0 accuracy (92.0\%) is not only better than NASNetMobile (91.7\%) but EfficientNetB0 footprint (4.6 M) is also smaller than NASNetMobile (4.8 M), we select EfficientNetB0 architecture for further experiments.

Given EfficientNetB0 backbone, we evaluate our proposed RSIC system applying three types of attention layers: SE, CBAM, and the proposed Multihead attention.
In this experiment, only thee feature maps which are extracted from the final three convolutional bocks (block 5 to block 7) in the EfficientNetB0 backbone are used. 
As Table~\ref{table:res02} shows, applying attention layers helps to improve the system performance by 0.1\%, 0.3\%, and 1.8\% with SE, CBAM, and the proposed Multihead attention, respectively.

As the proposed Multihead attention layer outperforms SE and CBAM layers, we then evaluate the proposed Multihead attention with different number of feature maps.
As the results are shown in Table~\ref{table:res03}, using three feature maps still achieves the best performance.
Regarding the model complexity, using the proposed Multihead attention layer with three feature maps increases the model footprint from 4.6 M to 9.4 M parameters.
To meet the constraints of maximum 20 MB of memory occupation, we apply the quantization technique which helps to reduce the model complexity to 9.4 MB (i.e. The quantization technique help to quantize a 32-bit floating point to 8-bit integer, then reduce the model footprint to 1/4 of the original footprint).
Notably, although the pruning techniques can help to significantly reduce a deep learning model to 1/10 of the original size~\cite{pajusco2020lightweight}, pruning parameters considered as zero still occupy the memorize of edge devices and cost the same computation as the non-pruning parameters.
Therefore, the pruning technique is not applied in this paper.

By using EfficientNetB0 as CNN-based backbone, the transfer learning, the proposed Multihead attention layer for three feature maps and the quantization technique, we achieve a low-complexity RISC model (9.4 MB).
We evaluate this model on NWPU-RESISC45 with two splitting settings as mentioned in Section~\ref{dataset} and compare with the state-of-the-art systems.
As Table \ref{table:res_11} shows, we can see that our results are very competitive compared with the state-of-the-art systems.
We achieve accuracy scores of 91.0\% and 93.8\% with training proportions of 10\% and 20\% respectively.
Compared with the system also using EfficientNetB0 in~\cite{sota_05}, our proposed RSIC not only outperforms but also presents a lower model footprint.
Our proposed system performs lower than 2\% compared with the best model using a Transformer based architecture~\cite{sota_13}. 
However, our model presents a significantly low memory occupation (9.4 M/9.4 MB) compared with the Transformer based model.

\begin{table}[t]
    \caption{Performance comparison of EfficientB0 with the transfer learning and different attention schemes on the benchmark NWPU-RESISC45 dataset using 20\%-80\% splitting settings.} 
    \centering
    \scalebox{1.0}{
    \begin{tabular}{|l|c|c|c|} 
        \hline 
        \textbf{Attention}  &\textbf{SE} &\textbf{CBAM} &\textbf{Proposed Multihead}   \\
	    \hline         
	            \textbf{Accuracy (\%)}  &92.1  &92.3 &93.8  \\
	            \textbf{Parameters (M)}  &10.4  &6.7 &9.4  \\
         \hline 
      \end{tabular}
    }
    \label{table:res02} 
\end{table}
\begin{table}[t]
    \caption{Performance (Acc.\%) comparison to the state-of-the-art systems on the benchmark NWPU-RESISC45 dataset with two splitting settings.} 
    \centering
    \scalebox{0.9}{

    \begin{tabular}{|l|c|c|} 
        \hline 
\textbf{Methods}                                       &\textbf{10\% training} &\textbf{20\% training} \\
	    \hline         
MG-CAP (Log-E) (55.99 M)~\cite{sota_01}                &89.4                   &91.7 \\ 
MG-CAP (Bilinear) (55.99 M)~\cite{sota_01}             &89.4                   &93.0 \\
MG-CAP (Sqrt-E) (55.99 M)~\cite{sota_01}               &90.8                   &93.0 \\ 
EfficientNet-B0-aux ($\approx 5.3 M$)~\cite{sota_05}   &90.0                   &92.9  \\
EfficientNet-B3-aux ($\approx 13 M$)~\cite{sota_05}    &91.1                   &93.8  \\
VGG-16 + MTL ($\approx$ 138.4 M)~\cite{sota_11}        &-                      &91.5 \\ 
ResNeXt-50 + MTL ($\approx$ 25 M)~\cite{sota_11}       &-                      &93.8 \\
ResNeXt-101 + MTL ($\approx$ 88.79 M)~\cite{sota_11}   &91.9                   &94.2 \\
SE-MDPMNet (5.17 M)~\cite{sota_10}                     &91.8                   &94.1 \\
LGRIN (4.63 M)~\cite{sota_12}                          &91.9                   &94.4 \\
Transformer (46.3 M)~\cite{sota_13}                    &93.1                   &95.6 \\ 
        \hline 
\textbf{Our systems (9.4 M / 9.4 MB)}                           &\textbf{91.0}          &\textbf{93.8} \\
         \hline 
      \end{tabular}
    }
    \label{table:res_11} 
\end{table}
\section{Conclusion}
This paper has presented a deep learning based model for remote sensing image classification (RSIC).
By conducting extensive experiments, we indicate that applying multiple techniques of transfer learning, Multihead attention on  multiple feature maps, and quantization to EfficientNetB0 based network architecture helps to achieve a high-performance and low-complexity RSIC system. 
The experimental results prove our proposed RSIC system competitive to the state-of-the-art systems and potential to apply on a wide range of edge devices.

\begin{acks}
    We would like to thank Ho Chi Minh City University of Technology (HCMUT), Vietnam National University Ho Chi Minh City (VNU-HCM) for the support of time and facilities for this study.
\end{acks}

\bibliographystyle{ACM-Reference-Format}
\bibliography{sample-base}


\begin{thebibliography}{65}


\ifx \showCODEN    \undefined \def \showCODEN     #1{\unskip}     \fi
\ifx \showDOI      \undefined \def \showDOI       #1{#1}\fi
\ifx \showISBNx    \undefined \def \showISBNx     #1{\unskip}     \fi
\ifx \showISBNxiii \undefined \def \showISBNxiii  #1{\unskip}     \fi
\ifx \showISSN     \undefined \def \showISSN      #1{\unskip}     \fi
\ifx \showLCCN     \undefined \def \showLCCN      #1{\unskip}     \fi
\ifx \shownote     \undefined \def \shownote      #1{#1}          \fi
\ifx \showarticletitle \undefined \def \showarticletitle #1{#1}   \fi
\ifx \showURL      \undefined \def \showURL       {\relax}        \fi
\providecommand\bibfield[2]{#2}
\providecommand\bibinfo[2]{#2}
\providecommand\natexlab[1]{#1}
\providecommand\showeprint[2][]{arXiv:#2}

\bibitem[Basu et~al\mbox{.}(2015)]%
        {basu2015deepsat}
\bibfield{author}{\bibinfo{person}{Saikat Basu}, \bibinfo{person}{Sangram
  Ganguly}, \bibinfo{person}{Supratik Mukhopadhyay}, \bibinfo{person}{Robert
  DiBiano}, \bibinfo{person}{Manohar Karki}, {and} \bibinfo{person}{Ramakrishna
  Nemani}.} \bibinfo{year}{2015}\natexlab{}.
\newblock \showarticletitle{Deepsat: a learning framework for satellite
  imagery}. In \bibinfo{booktitle}{\emph{Proceedings of the 23rd SIGSPATIAL
  international conference on advances in geographic information systems}}.
  \bibinfo{pages}{1--10}.
\newblock


\bibitem[Bazi et~al\mbox{.}(2019)]%
        {sota_05}
\bibfield{author}{\bibinfo{person}{Yakoub Bazi}, \bibinfo{person}{Mohamad~M
  Al~Rahhal}, \bibinfo{person}{Haikel Alhichri}, {and} \bibinfo{person}{Naif
  Alajlan}.} \bibinfo{year}{2019}\natexlab{}.
\newblock \showarticletitle{Simple yet effective fine-tuning of deep CNNs using
  an auxiliary classification loss for remote sensing scene classification}.
\newblock \bibinfo{journal}{\emph{Remote Sensing}} \bibinfo{volume}{11},
  \bibinfo{number}{24} (\bibinfo{year}{2019}), \bibinfo{pages}{2908}.
\newblock


\bibitem[Cheng et~al\mbox{.}(2017a)]%
        {cheng2017remote}
\bibfield{author}{\bibinfo{person}{Gong Cheng}, \bibinfo{person}{Junwei Han},
  {and} \bibinfo{person}{Xiaoqiang Lu}.} \bibinfo{year}{2017}\natexlab{a}.
\newblock \showarticletitle{Remote sensing image scene classification:
  Benchmark and state of the art}.
\newblock \bibinfo{journal}{\emph{Proc. IEEE}} \bibinfo{volume}{105},
  \bibinfo{number}{10} (\bibinfo{year}{2017}), \bibinfo{pages}{1865--1883}.
\newblock


\bibitem[Cheng et~al\mbox{.}(2017b)]%
        {nwpu_dataset}
\bibfield{author}{\bibinfo{person}{Gong Cheng}, \bibinfo{person}{Junwei Han},
  {and} \bibinfo{person}{Xiaoqiang Lu}.} \bibinfo{year}{2017}\natexlab{b}.
\newblock \showarticletitle{Remote Sensing Image Scene Classification:
  Benchmark and State of the Art}.
\newblock \bibinfo{journal}{\emph{Proc. IEEE}} \bibinfo{volume}{105},
  \bibinfo{number}{10} (\bibinfo{year}{2017}), \bibinfo{pages}{1865--1883}.
\newblock


\bibitem[Cheng et~al\mbox{.}(2014)]%
        {cheng2014multi}
\bibfield{author}{\bibinfo{person}{Gong Cheng}, \bibinfo{person}{Junwei Han},
  \bibinfo{person}{Peicheng Zhou}, {and} \bibinfo{person}{Lei Guo}.}
  \bibinfo{year}{2014}\natexlab{}.
\newblock \showarticletitle{Multi-class geospatial object detection and
  geographic image classification based on collection of part detectors}.
\newblock \bibinfo{journal}{\emph{ISPRS Journal of Photogrammetry and Remote
  Sensing}}  \bibinfo{volume}{98} (\bibinfo{year}{2014}),
  \bibinfo{pages}{119--132}.
\newblock


\bibitem[Chollet et~al\mbox{.}(2015)]%
        {keras_app}
\bibfield{author}{\bibinfo{person}{Fran\c{c}ois Chollet} {et~al\mbox{.}}}
  \bibinfo{year}{2015}\natexlab{}.
\newblock \bibinfo{title}{Keras}.
\newblock \bibinfo{howpublished}{\url{https://keras.io}}.
\newblock


\bibitem[Deng et~al\mbox{.}(2009)]%
        {imagenet_dataset}
\bibfield{author}{\bibinfo{person}{Jia Deng}, \bibinfo{person}{Wei Dong},
  \bibinfo{person}{Richard Socher}, \bibinfo{person}{Li-Jia Li},
  \bibinfo{person}{Kai Li}, {and} \bibinfo{person}{Li Fei-Fei}.}
  \bibinfo{year}{2009}\natexlab{}.
\newblock \showarticletitle{ImageNet: A large-scale hierarchical image
  database}. In \bibinfo{booktitle}{\emph{IEEE Conference on Computer Vision
  and Pattern Recognition}}. \bibinfo{pages}{248--255}.
\newblock


\bibitem[Du et~al\mbox{.}(2012)]%
        {du2012multiple}
\bibfield{author}{\bibinfo{person}{Peijun Du}, \bibinfo{person}{Junshi Xia},
  \bibinfo{person}{Wei Zhang}, \bibinfo{person}{Kun Tan}, \bibinfo{person}{Yi
  Liu}, {and} \bibinfo{person}{Sicong Liu}.} \bibinfo{year}{2012}\natexlab{}.
\newblock \showarticletitle{Multiple classifier system for remote sensing image
  classification: A review}.
\newblock \bibinfo{journal}{\emph{Sensors}} \bibinfo{volume}{12},
  \bibinfo{number}{4} (\bibinfo{year}{2012}), \bibinfo{pages}{4764--4792}.
\newblock


\bibitem[Elmannai et~al\mbox{.}(2016)]%
        {elmannai2016new}
\bibfield{author}{\bibinfo{person}{Hela Elmannai}, \bibinfo{person}{MohamedAnis
  Loghmari}, {and} \bibinfo{person}{Mohamed~Saber Naceur}.}
  \bibinfo{year}{2016}\natexlab{}.
\newblock \showarticletitle{A new classification approach based on source
  separation and feature extraction}. In \bibinfo{booktitle}{\emph{2016
  International Symposium on Signal, Image, Video and Communications (ISIVC)}}.
  IEEE, \bibinfo{pages}{137--141}.
\newblock


\bibitem[Elmannai et~al\mbox{.}(2013)]%
        {elmannai2013support}
\bibfield{author}{\bibinfo{person}{Hela Elmannai},
  \bibinfo{person}{Mohamed~Anis Loghmari}, {and} \bibinfo{person}{Mohamed~Saber
  Naceur}.} \bibinfo{year}{2013}\natexlab{}.
\newblock \showarticletitle{Support vector machine for remote sensing image
  classification}. In \bibinfo{booktitle}{\emph{Proceedings Engineering \&
  Technology}}, Vol.~\bibinfo{volume}{2}. \bibinfo{pages}{68--72}.
\newblock


\bibitem[Feng et~al\mbox{.}(2015)]%
        {feng2015uav}
\bibfield{author}{\bibinfo{person}{Quanlong Feng}, \bibinfo{person}{Jiantao
  Liu}, {and} \bibinfo{person}{Jianhua Gong}.} \bibinfo{year}{2015}\natexlab{}.
\newblock \showarticletitle{UAV remote sensing for urban vegetation mapping
  using random forest and texture analysis}.
\newblock \bibinfo{journal}{\emph{Remote sensing}} \bibinfo{volume}{7},
  \bibinfo{number}{1} (\bibinfo{year}{2015}), \bibinfo{pages}{1074--1094}.
\newblock


\bibitem[Guo et~al\mbox{.}(2019)]%
        {guo2019global}
\bibfield{author}{\bibinfo{person}{Yiyou Guo}, \bibinfo{person}{Jinsheng Ji},
  \bibinfo{person}{Xiankai Lu}, \bibinfo{person}{Hong Huo},
  \bibinfo{person}{Tao Fang}, {and} \bibinfo{person}{Deren Li}.}
  \bibinfo{year}{2019}\natexlab{}.
\newblock \showarticletitle{Global-local attention network for aerial scene
  classification}.
\newblock \bibinfo{journal}{\emph{IEEE Access}}  \bibinfo{volume}{7}
  (\bibinfo{year}{2019}), \bibinfo{pages}{67200--67212}.
\newblock


\bibitem[Hu et~al\mbox{.}(2015)]%
        {hu2015transferring}
\bibfield{author}{\bibinfo{person}{Fan Hu}, \bibinfo{person}{Gui-Song Xia},
  \bibinfo{person}{Jingwen Hu}, {and} \bibinfo{person}{Liangpei Zhang}.}
  \bibinfo{year}{2015}\natexlab{}.
\newblock \showarticletitle{Transferring deep convolutional neural networks for
  the scene classification of high-resolution remote sensing imagery}.
\newblock \bibinfo{journal}{\emph{Remote Sensing}} \bibinfo{volume}{7},
  \bibinfo{number}{11} (\bibinfo{year}{2015}), \bibinfo{pages}{14680--14707}.
\newblock


\bibitem[Hu et~al\mbox{.}(2018)]%
        {hu2018squeeze}
\bibfield{author}{\bibinfo{person}{Jie Hu}, \bibinfo{person}{Li Shen}, {and}
  \bibinfo{person}{Gang Sun}.} \bibinfo{year}{2018}\natexlab{}.
\newblock \showarticletitle{Squeeze-and-excitation networks}. In
  \bibinfo{booktitle}{\emph{Proceedings of the IEEE conference on computer
  vision and pattern recognition}}. \bibinfo{pages}{7132--7141}.
\newblock


\bibitem[Kingma and Ba(2015)]%
        {adam}
\bibfield{author}{\bibinfo{person}{Diederik~P. Kingma} {and}
  \bibinfo{person}{Jimmy Ba}.} \bibinfo{year}{2015}\natexlab{}.
\newblock \showarticletitle{Adam: A Method for Stochastic Optimization}.
\newblock \bibinfo{journal}{\emph{CoRR}}  \bibinfo{volume}{abs/1412.6980}
  (\bibinfo{year}{2015}).
\newblock


\bibitem[Kullback and Leibler(1951)]%
        {kl_loss}
\bibfield{author}{\bibinfo{person}{Solomon Kullback} {and}
  \bibinfo{person}{Richard~A Leibler}.} \bibinfo{year}{1951}\natexlab{}.
\newblock \showarticletitle{On information and sufficiency}.
\newblock \bibinfo{journal}{\emph{The annals of mathematical statistics}}
  \bibinfo{volume}{22}, \bibinfo{number}{1} (\bibinfo{year}{1951}),
  \bibinfo{pages}{79--86}.
\newblock


\bibitem[Li et~al\mbox{.}(2021)]%
        {li2021gated}
\bibfield{author}{\bibinfo{person}{Boyang Li}, \bibinfo{person}{Yulan Guo},
  \bibinfo{person}{Jungang Yang}, \bibinfo{person}{Longguang Wang},
  \bibinfo{person}{Yingqian Wang}, {and} \bibinfo{person}{Wei An}.}
  \bibinfo{year}{2021}\natexlab{}.
\newblock \showarticletitle{Gated recurrent multiattention network for VHR
  remote sensing image classification}.
\newblock \bibinfo{journal}{\emph{IEEE Transactions on Geoscience and Remote
  Sensing}}  \bibinfo{volume}{60} (\bibinfo{year}{2021}),
  \bibinfo{pages}{1--13}.
\newblock


\bibitem[Li et~al\mbox{.}(2020)]%
        {li2020augmentation}
\bibfield{author}{\bibinfo{person}{Fengpeng Li}, \bibinfo{person}{Ruyi Feng},
  \bibinfo{person}{Wei Han}, {and} \bibinfo{person}{Lizhe Wang}.}
  \bibinfo{year}{2020}\natexlab{}.
\newblock \showarticletitle{An augmentation attention mechanism for
  high-spatial-resolution remote sensing image scene classification}.
\newblock \bibinfo{journal}{\emph{IEEE Journal of Selected Topics in Applied
  Earth Observations and Remote Sensing}}  \bibinfo{volume}{13}
  (\bibinfo{year}{2020}), \bibinfo{pages}{3862--3878}.
\newblock


\bibitem[Mehmood et~al\mbox{.}(2022)]%
        {mehmood2022remote}
\bibfield{author}{\bibinfo{person}{Maryam Mehmood}, \bibinfo{person}{Ahsan
  Shahzad}, \bibinfo{person}{Bushra Zafar}, \bibinfo{person}{Amsa Shabbir},
  {and} \bibinfo{person}{Nouman Ali}.} \bibinfo{year}{2022}\natexlab{}.
\newblock \showarticletitle{Remote sensing image classification: A
  comprehensive review and applications}.
\newblock \bibinfo{journal}{\emph{Mathematical Problems in Engineering}}
  \bibinfo{volume}{2022} (\bibinfo{year}{2022}).
\newblock


\bibitem[Minetto et~al\mbox{.}(2019)]%
        {minetto2019hydra}
\bibfield{author}{\bibinfo{person}{Rodrigo Minetto},
  \bibinfo{person}{Maur{\'\i}cio~Pamplona Segundo}, {and}
  \bibinfo{person}{Sudeep Sarkar}.} \bibinfo{year}{2019}\natexlab{}.
\newblock \showarticletitle{Hydra: An ensemble of convolutional neural networks
  for geospatial land classification}.
\newblock \bibinfo{journal}{\emph{IEEE Transactions on Geoscience and Remote
  Sensing}} \bibinfo{volume}{57}, \bibinfo{number}{9} (\bibinfo{year}{2019}),
  \bibinfo{pages}{6530--6541}.
\newblock


\bibitem[Nair and Hinton(2010)]%
        {relu}
\bibfield{author}{\bibinfo{person}{Vinod Nair} {and}
  \bibinfo{person}{Geoffrey~E Hinton}.} \bibinfo{year}{2010}\natexlab{}.
\newblock \showarticletitle{Rectified linear units improve restricted boltzmann
  machines}. In \bibinfo{booktitle}{\emph{International Conference on Machine
  Learning (ICML)}}.
\newblock


\bibitem[Netzband et~al\mbox{.}(2007)]%
        {netzband2007applied}
\bibfield{author}{\bibinfo{person}{Maik Netzband}, \bibinfo{person}{William~L
  Stefanov}, {and} \bibinfo{person}{Charles Redman}.}
  \bibinfo{year}{2007}\natexlab{}.
\newblock \bibinfo{booktitle}{\emph{Applied remote sensing for urban planning,
  governance and sustainability}}.
\newblock \bibinfo{publisher}{Springer Science \& Business Media}.
\newblock


\bibitem[Nogueira et~al\mbox{.}(2017)]%
        {nogueira2017towards}
\bibfield{author}{\bibinfo{person}{Keiller Nogueira},
  \bibinfo{person}{Ot{\'a}vio~AB Penatti}, {and} \bibinfo{person}{Jefersson~A
  Dos~Santos}.} \bibinfo{year}{2017}\natexlab{}.
\newblock \showarticletitle{Towards better exploiting convolutional neural
  networks for remote sensing scene classification}.
\newblock \bibinfo{journal}{\emph{Pattern Recognition}}  \bibinfo{volume}{61}
  (\bibinfo{year}{2017}), \bibinfo{pages}{539--556}.
\newblock


\bibitem[Pajusco et~al\mbox{.}(2020)]%
        {pajusco2020lightweight}
\bibfield{author}{\bibinfo{person}{Nicolas Pajusco}, \bibinfo{person}{Richard
  Huang}, {and} \bibinfo{person}{Nicolas Farrugia}.}
  \bibinfo{year}{2020}\natexlab{}.
\newblock \showarticletitle{Lightweight Convolutional Neural Networks on
  Binaural Waveforms for Low Complexity Acoustic Scene Classification.}. In
  \bibinfo{booktitle}{\emph{DCASE}}. \bibinfo{pages}{135--139}.
\newblock


\bibitem[Pham et~al\mbox{.}(2022)]%
        {pham2022remote}
\bibfield{author}{\bibinfo{person}{Lam Pham}, \bibinfo{person}{Khoa Tran},
  \bibinfo{person}{Dat Ngo}, \bibinfo{person}{Jasmin Lampert}, {and}
  \bibinfo{person}{Alexander Schindler}.} \bibinfo{year}{2022}\natexlab{}.
\newblock \showarticletitle{Remote Sensing Image Classification using Transfer
  Learning and Attention Based Deep Neural Network}.
\newblock \bibinfo{journal}{\emph{arXiv preprint arXiv:2206.13392}}
  (\bibinfo{year}{2022}).
\newblock


\bibitem[Pires~de Lima and Marfurt(2019)]%
        {pires2019convolutional}
\bibfield{author}{\bibinfo{person}{Rafael Pires~de Lima} {and}
  \bibinfo{person}{Kurt Marfurt}.} \bibinfo{year}{2019}\natexlab{}.
\newblock \showarticletitle{Convolutional neural network for remote-sensing
  scene classification: Transfer learning analysis}.
\newblock \bibinfo{journal}{\emph{Remote Sensing}} \bibinfo{volume}{12},
  \bibinfo{number}{1} (\bibinfo{year}{2019}), \bibinfo{pages}{86}.
\newblock


\bibitem[Poursanidis and Chrysoulakis(2017)]%
        {poursanidis2017remote}
\bibfield{author}{\bibinfo{person}{Dimitris Poursanidis} {and}
  \bibinfo{person}{Nektarios Chrysoulakis}.} \bibinfo{year}{2017}\natexlab{}.
\newblock \showarticletitle{Remote Sensing, natural hazards and the
  contribution of ESA Sentinels missions}.
\newblock \bibinfo{journal}{\emph{Remote Sensing Applications: Society and
  Environment}}  \bibinfo{volume}{6} (\bibinfo{year}{2017}),
  \bibinfo{pages}{25--38}.
\newblock


\bibitem[Russakovsky et~al\mbox{.}(2015)]%
        {Imagenet}
\bibfield{author}{\bibinfo{person}{Olga Russakovsky}, \bibinfo{person}{Jia
  Deng}, \bibinfo{person}{Hao Su}, \bibinfo{person}{Jonathan Krause},
  \bibinfo{person}{Sanjeev Satheesh}, \bibinfo{person}{Sean Ma},
  \bibinfo{person}{Zhiheng Huang}, \bibinfo{person}{Andrej Karpathy},
  \bibinfo{person}{Aditya Khosla}, \bibinfo{person}{Michael Bernstein},
  \bibinfo{person}{Alexander~C. Berg}, {and} \bibinfo{person}{Li Fei-Fei}.}
  \bibinfo{year}{2015}\natexlab{}.
\newblock \showarticletitle{{ImageNet Large Scale Visual Recognition
  Challenge}}.
\newblock \bibinfo{journal}{\emph{International Journal of Computer Vision
  (IJCV)}} \bibinfo{number}{3} (\bibinfo{year}{2015}),
  \bibinfo{pages}{211--252}.
\newblock


\bibitem[Shabbir et~al\mbox{.}(2021)]%
        {shabbir2021satellite}
\bibfield{author}{\bibinfo{person}{Amsa Shabbir}, \bibinfo{person}{Nouman Ali},
  \bibinfo{person}{Jameel Ahmed}, \bibinfo{person}{Bushra Zafar},
  \bibinfo{person}{Aqsa Rasheed}, \bibinfo{person}{Muhammad Sajid},
  \bibinfo{person}{Afzal Ahmed}, {and} \bibinfo{person}{Saadat~Hanif Dar}.}
  \bibinfo{year}{2021}\natexlab{}.
\newblock \showarticletitle{Satellite and scene image classification based on
  transfer learning and fine tuning of ResNet50}.
\newblock \bibinfo{journal}{\emph{Mathematical Problems in Engineering}}
  \bibinfo{volume}{2021} (\bibinfo{year}{2021}).
\newblock


\bibitem[Shorten and Khoshgoftaar(2019)]%
        {rotation_aug}
\bibfield{author}{\bibinfo{person}{Connor Shorten} {and}
  \bibinfo{person}{Taghi~M Khoshgoftaar}.} \bibinfo{year}{2019}\natexlab{}.
\newblock \showarticletitle{A survey on image data augmentation for deep
  learning}.
\newblock \bibinfo{journal}{\emph{Journal of big data}} \bibinfo{volume}{6},
  \bibinfo{number}{1} (\bibinfo{year}{2019}), \bibinfo{pages}{1--48}.
\newblock


\bibitem[Sridharan and Cheriyadat(2014)]%
        {sridharan2014bag}
\bibfield{author}{\bibinfo{person}{Harini Sridharan} {and}
  \bibinfo{person}{Anil Cheriyadat}.} \bibinfo{year}{2014}\natexlab{}.
\newblock \showarticletitle{Bag of lines (BoL) for improved aerial scene
  representation}.
\newblock \bibinfo{journal}{\emph{IEEE Geoscience and Remote Sensing Letters}}
  \bibinfo{volume}{12}, \bibinfo{number}{3} (\bibinfo{year}{2014}),
  \bibinfo{pages}{676--680}.
\newblock


\bibitem[Srivastava et~al\mbox{.}(2014)]%
        {dropout}
\bibfield{author}{\bibinfo{person}{Nitish Srivastava},
  \bibinfo{person}{Geoffrey Hinton}, \bibinfo{person}{Alex Krizhevsky},
  \bibinfo{person}{Ilya Sutskever}, {and} \bibinfo{person}{Ruslan
  Salakhutdinov}.} \bibinfo{year}{2014}\natexlab{}.
\newblock \showarticletitle{Dropout: a simple way to prevent neural networks
  from overfitting}.
\newblock \bibinfo{journal}{\emph{The {J}ournal of {M}achine {L}earning
  {R}esearch}} \bibinfo{volume}{15}, \bibinfo{number}{1}
  (\bibinfo{year}{2014}), \bibinfo{pages}{1929--1958}.
\newblock


\bibitem[Sun et~al\mbox{.}(2021)]%
        {sun2021mind}
\bibfield{author}{\bibinfo{person}{Zhichuang Sun}, \bibinfo{person}{Ruimin
  Sun}, \bibinfo{person}{Long Lu}, {and} \bibinfo{person}{Alan Mislove}.}
  \bibinfo{year}{2021}\natexlab{}.
\newblock \showarticletitle{Mind your weight (s): A large-scale study on
  insufficient machine learning model protection in mobile apps}. In
  \bibinfo{booktitle}{\emph{30th USENIX Security Symposium (USENIX Security
  21)}}. \bibinfo{pages}{1955--1972}.
\newblock


\bibitem[Takahashi et~al\mbox{.}(2020)]%
        {spec_crop}
\bibfield{author}{\bibinfo{person}{Ryo Takahashi}, \bibinfo{person}{Takashi
  Matsubara}, {and} \bibinfo{person}{Kuniaki Uehara}.}
  \bibinfo{year}{2020}\natexlab{}.
\newblock \showarticletitle{Data Augmentation Using Random Image Cropping and
  Patching for Deep CNNs}.
\newblock \bibinfo{journal}{\emph{IEEE Transactions on Circuits and Systems for
  Video Technology}} \bibinfo{volume}{30}, \bibinfo{number}{9}
  (\bibinfo{year}{2020}), \bibinfo{pages}{2917--2931}.
\newblock


\bibitem[Tan and Le(2019)]%
        {tan2019efficientnet}
\bibfield{author}{\bibinfo{person}{Mingxing Tan} {and} \bibinfo{person}{Quoc
  Le}.} \bibinfo{year}{2019}\natexlab{}.
\newblock \showarticletitle{Efficientnet: Rethinking model scaling for
  convolutional neural networks}. In \bibinfo{booktitle}{\emph{International
  conference on machine learning}}. PMLR, \bibinfo{pages}{6105--6114}.
\newblock


\bibitem[Thapa and Murayama(2009)]%
        {thapa2009urban}
\bibfield{author}{\bibinfo{person}{Rajesh~Bahadur Thapa} {and}
  \bibinfo{person}{Yuji Murayama}.} \bibinfo{year}{2009}\natexlab{}.
\newblock \showarticletitle{Urban mapping, accuracy, \& image classification: A
  comparison of multiple approaches in Tsukuba City, Japan}.
\newblock \bibinfo{journal}{\emph{Applied geography}} \bibinfo{volume}{29},
  \bibinfo{number}{1} (\bibinfo{year}{2009}), \bibinfo{pages}{135--144}.
\newblock


\bibitem[Tokozume et~al\mbox{.}(2018)]%
        {mixup2}
\bibfield{author}{\bibinfo{person}{Y. Tokozume}, \bibinfo{person}{Y. Ushiku},
  {and} \bibinfo{person}{T. Harada}.} \bibinfo{year}{2018}\natexlab{}.
\newblock \showarticletitle{Learning from between-class examples for deep sound
  recognition}.
\newblock \bibinfo{journal}{\emph{in ICLR}} (\bibinfo{year}{2018}).
\newblock


\bibitem[Tong et~al\mbox{.}(2020)]%
        {tong2020channel}
\bibfield{author}{\bibinfo{person}{Wei Tong}, \bibinfo{person}{Weitao Chen},
  \bibinfo{person}{Wei Han}, \bibinfo{person}{Xianju Li}, {and}
  \bibinfo{person}{Lizhe Wang}.} \bibinfo{year}{2020}\natexlab{}.
\newblock \showarticletitle{Channel-attention-based DenseNet network for remote
  sensing image scene classification}.
\newblock \bibinfo{journal}{\emph{IEEE Journal of Selected Topics in Applied
  Earth Observations and Remote Sensing}}  \bibinfo{volume}{13}
  (\bibinfo{year}{2020}), \bibinfo{pages}{4121--4132}.
\newblock


\bibitem[Van~Westen(2013)]%
        {van2013remote}
\bibfield{author}{\bibinfo{person}{Cees~J Van~Westen}.}
  \bibinfo{year}{2013}\natexlab{}.
\newblock \showarticletitle{Remote sensing and GIS for natural hazards
  assessment and disaster risk management}.
\newblock \bibinfo{journal}{\emph{Treatise on geomorphology}}
  \bibinfo{volume}{3} (\bibinfo{year}{2013}), \bibinfo{pages}{259--298}.
\newblock


\bibitem[Vaswani et~al\mbox{.}(2017)]%
        {vaswani2017attention}
\bibfield{author}{\bibinfo{person}{Ashish Vaswani}, \bibinfo{person}{Noam
  Shazeer}, \bibinfo{person}{Niki Parmar}, \bibinfo{person}{Jakob Uszkoreit},
  \bibinfo{person}{Llion Jones}, \bibinfo{person}{Aidan~N Gomez},
  \bibinfo{person}{{\L}ukasz Kaiser}, {and} \bibinfo{person}{Illia
  Polosukhin}.} \bibinfo{year}{2017}\natexlab{}.
\newblock \showarticletitle{Attention is all you need}.
\newblock \bibinfo{journal}{\emph{Advances in neural information processing
  systems}}  \bibinfo{volume}{30} (\bibinfo{year}{2017}).
\newblock


\bibitem[Wang et~al\mbox{.}(2022)]%
        {wang2022empirical}
\bibfield{author}{\bibinfo{person}{Di Wang}, \bibinfo{person}{Jing Zhang},
  \bibinfo{person}{Bo Du}, \bibinfo{person}{Gui-Song Xia}, {and}
  \bibinfo{person}{Dacheng Tao}.} \bibinfo{year}{2022}\natexlab{}.
\newblock \showarticletitle{An Empirical Study of Remote Sensing Pretraining}.
\newblock \bibinfo{journal}{\emph{IEEE Transactions on Geoscience and Remote
  Sensing}} (\bibinfo{year}{2022}).
\newblock


\bibitem[Wang et~al\mbox{.}(2018)]%
        {wang2018scene}
\bibfield{author}{\bibinfo{person}{Qi Wang}, \bibinfo{person}{Shaoteng Liu},
  \bibinfo{person}{Jocelyn Chanussot}, {and} \bibinfo{person}{Xuelong Li}.}
  \bibinfo{year}{2018}\natexlab{}.
\newblock \showarticletitle{Scene classification with recurrent attention of
  VHR remote sensing images}.
\newblock \bibinfo{journal}{\emph{IEEE Transactions on Geoscience and Remote
  Sensing}} \bibinfo{volume}{57}, \bibinfo{number}{2} (\bibinfo{year}{2018}),
  \bibinfo{pages}{1155--1167}.
\newblock


\bibitem[Wang et~al\mbox{.}(2020)]%
        {sota_01}
\bibfield{author}{\bibinfo{person}{Shidong Wang}, \bibinfo{person}{Yu Guan},
  {and} \bibinfo{person}{Ling Shao}.} \bibinfo{year}{2020}\natexlab{}.
\newblock \showarticletitle{Multi-granularity canonical appearance pooling for
  remote sensing scene classification}.
\newblock \bibinfo{journal}{\emph{IEEE Transactions on Image Processing}}
  \bibinfo{volume}{29} (\bibinfo{year}{2020}), \bibinfo{pages}{5396--5407}.
\newblock


\bibitem[Weiss et~al\mbox{.}(2016)]%
        {survey_transfer}
\bibfield{author}{\bibinfo{person}{Karl Weiss}, \bibinfo{person}{Taghi~M
  Khoshgoftaar}, {and} \bibinfo{person}{DingDing Wang}.}
  \bibinfo{year}{2016}\natexlab{}.
\newblock \showarticletitle{A survey of transfer learning}.
\newblock \bibinfo{journal}{\emph{Journal of Big data}} \bibinfo{volume}{3},
  \bibinfo{number}{1} (\bibinfo{year}{2016}), \bibinfo{pages}{1--40}.
\newblock


\bibitem[Woo et~al\mbox{.}(2018)]%
        {woo2018cbam}
\bibfield{author}{\bibinfo{person}{Sanghyun Woo}, \bibinfo{person}{Jongchan
  Park}, \bibinfo{person}{Joon-Young Lee}, {and} \bibinfo{person}{In~So
  Kweon}.} \bibinfo{year}{2018}\natexlab{}.
\newblock \showarticletitle{Cbam: Convolutional block attention module}. In
  \bibinfo{booktitle}{\emph{Proceedings of the European conference on computer
  vision (ECCV)}}. \bibinfo{pages}{3--19}.
\newblock


\bibitem[Xia et~al\mbox{.}(2017a)]%
        {xia2017aid}
\bibfield{author}{\bibinfo{person}{Gui-Song Xia}, \bibinfo{person}{Jingwen Hu},
  \bibinfo{person}{Fan Hu}, \bibinfo{person}{Baoguang Shi},
  \bibinfo{person}{Xiang Bai}, \bibinfo{person}{Yanfei Zhong},
  \bibinfo{person}{Liangpei Zhang}, {and} \bibinfo{person}{Xiaoqiang Lu}.}
  \bibinfo{year}{2017}\natexlab{a}.
\newblock \showarticletitle{AID: A benchmark data set for performance
  evaluation of aerial scene classification}.
\newblock \bibinfo{journal}{\emph{IEEE Transactions on Geoscience and Remote
  Sensing}} \bibinfo{volume}{55}, \bibinfo{number}{7} (\bibinfo{year}{2017}),
  \bibinfo{pages}{3965--3981}.
\newblock


\bibitem[Xia et~al\mbox{.}(2017b)]%
        {aid_dataset}
\bibfield{author}{\bibinfo{person}{Gui-Song Xia}, \bibinfo{person}{Jingwen Hu},
  \bibinfo{person}{Fan Hu}, \bibinfo{person}{Baoguang Shi},
  \bibinfo{person}{Xiang Bai}, \bibinfo{person}{Yanfei Zhong},
  \bibinfo{person}{Liangpei Zhang}, {and} \bibinfo{person}{Xiaoqiang Lu}.}
  \bibinfo{year}{2017}\natexlab{b}.
\newblock \showarticletitle{AID: A benchmark data set for performance
  evaluation of aerial scene classification}.
\newblock \bibinfo{journal}{\emph{IEEE Transactions on Geoscience and Remote
  Sensing}} \bibinfo{volume}{55}, \bibinfo{number}{7} (\bibinfo{year}{2017}),
  \bibinfo{pages}{3965--3981}.
\newblock


\bibitem[Xia et~al\mbox{.}(2010)]%
        {Xia2010WHURS19}
\bibfield{author}{\bibinfo{person}{Gui-Song Xia}, \bibinfo{person}{Wen Yang},
  \bibinfo{person}{Julie Delon}, \bibinfo{person}{Yann Gousseau},
  \bibinfo{person}{Hong Sun}, {and} \bibinfo{person}{Henri MaÎtre}.}
  \bibinfo{year}{2010}\natexlab{}.
\newblock \showarticletitle{Structural high-resolution satellite image
  indexing}.
\newblock \bibinfo{journal}{\emph{Symposium: 100 Years ISPRS - Advancing Remote
  Sensing Science}}.
\newblock


\bibitem[Xu et~al\mbox{.}(2021)]%
        {sota_12}
\bibfield{author}{\bibinfo{person}{Chengjun Xu}, \bibinfo{person}{Guobin Zhu},
  {and} \bibinfo{person}{Jingqian Shu}.} \bibinfo{year}{2021}\natexlab{}.
\newblock \showarticletitle{A lightweight and robust lie group-convolutional
  neural networks joint representation for remote sensing scene
  classification}.
\newblock \bibinfo{journal}{\emph{IEEE Transactions on Geoscience and Remote
  Sensing}}  \bibinfo{volume}{60} (\bibinfo{year}{2021}),
  \bibinfo{pages}{1--15}.
\newblock


\bibitem[Xu et~al\mbox{.}(2018)]%
        {mixup1}
\bibfield{author}{\bibinfo{person}{Kele Xu}, \bibinfo{person}{Dawei Feng},
  \bibinfo{person}{Haibo Mi}, \bibinfo{person}{Boqing Zhu},
  \bibinfo{person}{Dezhi Wang}, \bibinfo{person}{Lilun Zhang},
  \bibinfo{person}{Hengxing Cai}, {and} \bibinfo{person}{Shuwen Liu}.}
  \bibinfo{year}{2018}\natexlab{}.
\newblock \showarticletitle{Mixup-based acoustic scene classification using
  multi-channel convolutional neural network}. In
  \bibinfo{booktitle}{\emph{Pacific Rim Conference on Multimedia}}.
  \bibinfo{pages}{14--23}.
\newblock


\bibitem[Yang and Newsam(2008)]%
        {yang2008comparing}
\bibfield{author}{\bibinfo{person}{Yi Yang} {and} \bibinfo{person}{Shawn
  Newsam}.} \bibinfo{year}{2008}\natexlab{}.
\newblock \showarticletitle{Comparing SIFT descriptors and Gabor texture
  features for classification of remote sensed imagery}. In
  \bibinfo{booktitle}{\emph{2008 15th IEEE international conference on image
  processing}}. IEEE, \bibinfo{pages}{1852--1855}.
\newblock


\bibitem[Yang and Newsam(2010)]%
        {yang2010bag}
\bibfield{author}{\bibinfo{person}{Yi Yang} {and} \bibinfo{person}{Shawn
  Newsam}.} \bibinfo{year}{2010}\natexlab{}.
\newblock \showarticletitle{Bag-of-visual-words and spatial extensions for
  land-use classification}. In \bibinfo{booktitle}{\emph{Proceedings of the
  18th SIGSPATIAL international conference on advances in geographic
  information systems}}. \bibinfo{pages}{270--279}.
\newblock


\bibitem[Ye et~al\mbox{.}(2021)]%
        {ye2021lightweight}
\bibfield{author}{\bibinfo{person}{Mu Ye}, \bibinfo{person}{Ni Ruiwen},
  \bibinfo{person}{Zhang Chang}, \bibinfo{person}{Gong He}, \bibinfo{person}{Hu
  Tianli}, \bibinfo{person}{Li Shijun}, \bibinfo{person}{Sun Yu},
  \bibinfo{person}{Zhang Tong}, {and} \bibinfo{person}{Guo Ying}.}
  \bibinfo{year}{2021}\natexlab{}.
\newblock \showarticletitle{A Lightweight Model of VGG-16 for Remote Sensing
  Image Classification}.
\newblock \bibinfo{journal}{\emph{IEEE Journal of Selected Topics in Applied
  Earth Observations and Remote Sensing}}  \bibinfo{volume}{14}
  (\bibinfo{year}{2021}), \bibinfo{pages}{6916--6922}.
\newblock


\bibitem[Zhang et~al\mbox{.}(2019)]%
        {sota_10}
\bibfield{author}{\bibinfo{person}{Bin Zhang}, \bibinfo{person}{Yongjun Zhang},
  {and} \bibinfo{person}{Shugen Wang}.} \bibinfo{year}{2019}\natexlab{}.
\newblock \showarticletitle{A lightweight and discriminative model for remote
  sensing scene classification with multidilation pooling module}.
\newblock \bibinfo{journal}{\emph{IEEE Journal of Selected Topics in Applied
  Earth Observations and Remote Sensing}} \bibinfo{volume}{12},
  \bibinfo{number}{8} (\bibinfo{year}{2019}), \bibinfo{pages}{2636--2653}.
\newblock


\bibitem[Zhang et~al\mbox{.}(2020)]%
        {zhang2020transfer}
\bibfield{author}{\bibinfo{person}{Deyuan Zhang}, \bibinfo{person}{Zhenghong
  Liu}, {and} \bibinfo{person}{Xiangbin Shi}.} \bibinfo{year}{2020}\natexlab{}.
\newblock \showarticletitle{Transfer learning on EfficientNet for remote
  sensing image classification}. In \bibinfo{booktitle}{\emph{2020 5th
  International Conference on Mechanical, Control and Computer Engineering
  (ICMCCE)}}. IEEE, \bibinfo{pages}{2255--2258}.
\newblock


\bibitem[Zhang et~al\mbox{.}(2021b)]%
        {zhang2021trs}
\bibfield{author}{\bibinfo{person}{Jianrong Zhang}, \bibinfo{person}{Hongwei
  Zhao}, {and} \bibinfo{person}{Jiao Li}.} \bibinfo{year}{2021}\natexlab{b}.
\newblock \showarticletitle{TRS: Transformers for remote sensing scene
  classification}.
\newblock \bibinfo{journal}{\emph{Remote Sensing}} \bibinfo{volume}{13},
  \bibinfo{number}{20} (\bibinfo{year}{2021}), \bibinfo{pages}{4143}.
\newblock


\bibitem[Zhang et~al\mbox{.}(2021c)]%
        {sota_13}
\bibfield{author}{\bibinfo{person}{Jianrong Zhang}, \bibinfo{person}{Hongwei
  Zhao}, {and} \bibinfo{person}{Jiao Li}.} \bibinfo{year}{2021}\natexlab{c}.
\newblock \showarticletitle{TRS: Transformers for Remote Sensing Scene
  Classification}.
\newblock \bibinfo{journal}{\emph{Remote Sensing}} \bibinfo{volume}{13},
  \bibinfo{number}{20} (\bibinfo{year}{2021}), \bibinfo{pages}{4143}.
\newblock


\bibitem[Zhang et~al\mbox{.}(2021a)]%
        {zhang2021best}
\bibfield{author}{\bibinfo{person}{Xinqi Zhang}, \bibinfo{person}{Weining An},
  \bibinfo{person}{Jinggong Sun}, \bibinfo{person}{Hang Wu},
  \bibinfo{person}{Wenchang Zhang}, {and} \bibinfo{person}{Yaohua Du}.}
  \bibinfo{year}{2021}\natexlab{a}.
\newblock \showarticletitle{Best representation branch model for remote sensing
  image scene classification}.
\newblock \bibinfo{journal}{\emph{IEEE Journal of Selected Topics in Applied
  Earth Observations and Remote Sensing}}  \bibinfo{volume}{14}
  (\bibinfo{year}{2021}), \bibinfo{pages}{9768--9780}.
\newblock


\bibitem[Zhao et~al\mbox{.}(2015)]%
        {zhao2015dirichlet}
\bibfield{author}{\bibinfo{person}{Bei Zhao}, \bibinfo{person}{Yanfei Zhong},
  \bibinfo{person}{Gui-Song Xia}, {and} \bibinfo{person}{Liangpei Zhang}.}
  \bibinfo{year}{2015}\natexlab{}.
\newblock \showarticletitle{Dirichlet-derived multiple topic scene
  classification model for high spatial resolution remote sensing imagery}.
\newblock \bibinfo{journal}{\emph{IEEE Transactions on Geoscience and Remote
  Sensing}} \bibinfo{volume}{54}, \bibinfo{number}{4} (\bibinfo{year}{2015}),
  \bibinfo{pages}{2108--2123}.
\newblock


\bibitem[Zhao et~al\mbox{.}(2022)]%
        {Zhao_2022}
\bibfield{author}{\bibinfo{person}{Qi Zhao}, \bibinfo{person}{Yujing Ma},
  \bibinfo{person}{Shuchang Lyu}, {and} \bibinfo{person}{Lijiang Chen}.}
  \bibinfo{year}{2022}\natexlab{}.
\newblock \showarticletitle{Embedded Self-Distillation in Compact Multibranch
  Ensemble Network for Remote Sensing Scene Classification}.
\newblock \bibinfo{journal}{\emph{{IEEE} Transactions on Geoscience and Remote
  Sensing}}  \bibinfo{volume}{60} (\bibinfo{year}{2022}),
  \bibinfo{pages}{1--15}.
\newblock
\urldef\tempurl%
\url{https://doi.org/10.1109/tgrs.2021.3126770}
\showDOI{\tempurl}


\bibitem[Zhao et~al\mbox{.}(2020a)]%
        {zhao2020remote}
\bibfield{author}{\bibinfo{person}{Zhicheng Zhao}, \bibinfo{person}{Jiaqi Li},
  \bibinfo{person}{Ze Luo}, \bibinfo{person}{Jian Li}, {and}
  \bibinfo{person}{Can Chen}.} \bibinfo{year}{2020}\natexlab{a}.
\newblock \showarticletitle{Remote sensing image scene classification based on
  an enhanced attention module}.
\newblock \bibinfo{journal}{\emph{IEEE Geoscience and Remote Sensing Letters}}
  \bibinfo{volume}{18}, \bibinfo{number}{11} (\bibinfo{year}{2020}),
  \bibinfo{pages}{1926--1930}.
\newblock


\bibitem[Zhao et~al\mbox{.}(2020b)]%
        {sota_11}
\bibfield{author}{\bibinfo{person}{Zhicheng Zhao}, \bibinfo{person}{Ze Luo},
  \bibinfo{person}{Jian Li}, \bibinfo{person}{Can Chen}, {and}
  \bibinfo{person}{Yingchao Piao}.} \bibinfo{year}{2020}\natexlab{b}.
\newblock \showarticletitle{When self-supervised learning meets scene
  classification: Remote sensing scene classification based on a multitask
  learning framework}.
\newblock \bibinfo{journal}{\emph{Remote Sensing}} \bibinfo{volume}{12},
  \bibinfo{number}{20} (\bibinfo{year}{2020}), \bibinfo{pages}{3276}.
\newblock


\bibitem[Zheng et~al\mbox{.}(2008)]%
        {zheng2008k}
\bibfield{author}{\bibinfo{person}{Jian Zheng}, \bibinfo{person}{Zhanzhong
  Cui}, \bibinfo{person}{Anfei Liu}, {and} \bibinfo{person}{Yu Jia}.}
  \bibinfo{year}{2008}\natexlab{}.
\newblock \showarticletitle{A K-means remote sensing image classification
  method based on AdaBoost}. In \bibinfo{booktitle}{\emph{2008 Fourth
  International Conference on Natural Computation}}, Vol.~\bibinfo{volume}{4}.
  IEEE, \bibinfo{pages}{27--32}.
\newblock


\bibitem[Zhuang et~al\mbox{.}(2020)]%
        {survey_transfer_02}
\bibfield{author}{\bibinfo{person}{Fuzhen Zhuang}, \bibinfo{person}{Zhiyuan
  Qi}, \bibinfo{person}{Keyu Duan}, \bibinfo{person}{Dongbo Xi},
  \bibinfo{person}{Yongchun Zhu}, \bibinfo{person}{Hengshu Zhu},
  \bibinfo{person}{Hui Xiong}, {and} \bibinfo{person}{Qing He}.}
  \bibinfo{year}{2020}\natexlab{}.
\newblock \showarticletitle{A comprehensive survey on transfer learning}.
\newblock \bibinfo{journal}{\emph{Proc. IEEE}} \bibinfo{volume}{109},
  \bibinfo{number}{1} (\bibinfo{year}{2020}), \bibinfo{pages}{43--76}.
\newblock


\bibitem[Zou et~al\mbox{.}(2015)]%
        {zou2015deep}
\bibfield{author}{\bibinfo{person}{Qin Zou}, \bibinfo{person}{Lihao Ni},
  \bibinfo{person}{Tong Zhang}, {and} \bibinfo{person}{Qian Wang}.}
  \bibinfo{year}{2015}\natexlab{}.
\newblock \showarticletitle{Deep learning based feature selection for remote
  sensing scene classification}.
\newblock \bibinfo{journal}{\emph{IEEE Geoscience and Remote Sensing Letters}}
  \bibinfo{volume}{12}, \bibinfo{number}{11} (\bibinfo{year}{2015}),
  \bibinfo{pages}{2321--2325}.
\newblock


\end{thebibliography}

\end{document}